\documentclass[11pt,a4paper]{article}
\usepackage[hyperref]{eacl2021}
\usepackage{times}
\usepackage{latexsym}

\usepackage{microtype}
\usepackage{times}  
\usepackage{helvet} 
\usepackage{courier}  
\usepackage{natbib}  
\usepackage{caption} 
\usepackage[utf8]{inputenc}
\usepackage[english]{babel}

\usepackage{mathtools}
\DeclareMathOperator*{\argmax}{arg\,max}

\usepackage{cleveref}
\crefformat{section}{\S#2#1#3} 
\crefformat{subsection}{\S#2#1#3}
\crefformat{subsubsection}{\S#2#1#3}

\makeatletter
\def\hlinewd#1{%
\noalign{\ifnum0=`}\fi\hrule \@height #1 %
\futurelet\reserved@a\@xhline}
\makeatother

\usepackage{algorithm}

\usepackage{multirow}

\usepackage{amsmath}
\usepackage{subfig}
\usepackage{amsfonts}
\usepackage{algorithmicx}
\usepackage{algpseudocode}
\usepackage{multirow, makecell,booktabs,multicol}
\usepackage{mathtools}
\usepackage{latexsym}
\usepackage{url}
\usepackage{color}
\usepackage{makecell}
\usepackage{relsize}

\usepackage{cleveref}
\usepackage{flexisym}
\usepackage{arydshln}

\aclfinalcopy 

\title{Non-Autoregressive Text Generation with Pre-trained Language Models}

\author{Yixuan Su$^{\diamondsuit}$~\quad Deng Cai$^{\heartsuit}$\quad  Yan Wang$^{\spadesuit}$\quad David Vandyke$^{\clubsuit}$\quad \\
\textbf{Simon Baker}$^{\diamondsuit}$\quad 
\textbf{Piji Li}$^{\spadesuit}$\quad \textbf{Nigel Collier}$^\diamondsuit$\quad\\
$^\diamondsuit$Language Technology Lab, University of Cambridge \\
$^\heartsuit$The Chinese University of Hong Kong \\
$^\spadesuit$Tencent AI Lab \\
$^\clubsuit$Apple\\
{\tt \{ys484,sb895,nhc30\}@cam.ac.uk}\\ {\tt thisisjcykcd@gmail.com}{\tt ,} {\tt dvandyke@apple.com}\\ {\tt \{brandenwang,pijili\}@tencent.com}
}

\date{}

\begin{document}
\maketitle
\begin{abstract}
Non-autoregressive generation (NAG) has recently attracted great attention due to its fast inference speed. However, the generation quality of existing NAG models still lags behind their autoregressive counterparts. In this work, we show that BERT can be employed as the backbone of a NAG model to greatly improve performance. Additionally, we devise mechanisms to alleviate the two common problems of vanilla NAG models: the inflexibility of prefixed output length and the conditional independence of individual token predictions. Lastly, to further increase the speed advantage of the proposed model, we propose a new decoding strategy, \textit{ratio-first}, for applications where the output lengths can be approximately estimated beforehand. For a comprehensive evaluation, we test the proposed model on three text generation tasks, including text summarization, sentence compression and machine translation. Experimental results show that our model significantly outperforms existing non-autoregressive baselines and achieves competitive performance with many strong autoregressive models. In addition, we also conduct extensive analysis experiments to reveal the effect of each proposed component.\footnote{All related code, data, and models can be found in https://github.com/yxuansu/NAG-BERT.}
\end{abstract}

\section{Introduction}
Autoregressive generation (AG) models achieve state-of-the-art performance on a wide range of text generation tasks, such as machine translation \cite{DBLP:conf/nips/VaswaniSPUJGKP17} and text summarization \cite{DBLP:conf/emnlp/RushCW15}. Such models generate a token sequence in a left-to-right, token-by-token fashion. The prediction for the next token is conditioned on all previously generated tokens. This characteristic makes it impossible to parallelize the computational overhead for token predictions in different positions, which leads to a relatively high latency in inference. On the other hand, non-autoregressive generation (NAG) models \cite{DBLP:journals/corr/abs-1711-02281} have emerged as a promising alternative due to their fast inference speed. NAG models omit the sequential dependencies within the output-side sequence and predict tokens in all positions simultaneously once the output length has been determined beforehand. While NAG models enjoy full parallelism and faster inference, the generation quality of NAG models often lags behind their autoregressive counterparts.

In this work, we explore the potential of large-scale pre-trained language models for improving the performance of non-autoregressive generation. Specifically, we utilize BERT \cite{DBLP:conf/naacl/DevlinCLT19} as the backbone for NAG modelling and extend the architecture of BERT with a CRF output layer \cite{DBLP:conf/icml/LaffertyMP01,DBLP:conf/nips/SunLWHLD19} for better capturing the output-side dependencies. 

In addition, we analyze two significant limitations that NAG models currently suffer from: (1) the inflexibility of prefixed output length, and (2) the conditional independence of individual token predictions. Accordingly, we devise two solutions to these two problems.

First, prior NAG models require the output length to be determined before token generation, thus an extra module for output length prediction is always required. Nevertheless, the most likely length from the prediction module is not necessarily the best-suited one for the token generation model. To this end, previous works \cite{DBLP:journals/corr/abs-1711-02281,DBLP:conf/emnlp/MaZLNH19} usually rely on length-parallel decoding (LPD) \cite{DBLP:conf/acl/WeiWZLS19} for performance enhancement; that is, generating and re-ranking the results from different output length candidates. In this work, we propose a simple and elegant decoding mechanism that lets the model determine the output length on-the-fly. Specifically, our model dynamically adjusts the output sequence length via emitting an \texttt{[eos]} token at any output position to indicate the ending of the generated sequence. Therefore, we can avoid the additional efforts of output length prediction and results re-ranking.

Second, most existing NAG models assume the token predictions in different positions are conditionally independent. As a consequence, they often tend to generate results that are ungrammatical with repetitions \cite{DBLP:conf/aaai/WangTHQZL19}. To alleviate this problem, we propose a context-aware learning objective which impels the model to output different tokens at adjacent positions, thereby reducing the possibility of repetitive generation.

Furthermore, for tasks like text summarization, the output sequence (summary) is known to be shorter than the source sequence (article). In such cases, to further improve the model's inference efficiency, we introduce a new \textit{ratio-first} decoding strategy. Specifically, instead of performing inference on all source-side hidden states, ratio-first generates the result only based on a subset of source hidden states. The subset size is jointly determined by the source length $T$ and a predefined ratio $\alpha$ that is set based on our prior knowledge from the data statistics. In the experiments, we show that ratio-first can significantly improve the inference speed while maintaining the generation quality. 

We evaluate the proposed model on three typical text generation tasks, including text summarization, sentence compression and machine translation. Experimental results show that our model significantly outperforms many strong non-autoregressive baselines, and even performs competitively with several strong autoregressive models. In addition, we conduct extensive analysis experiments to study the effect of individual proposed components. 

In summary, our contributions are: 
(1) We propose a novel framework that utilizes BERT for text generation under the non-autoregressive generation paradigm; (2) We propose a decoding mechanism that allows the model to dynamically determine the output length, and a new context-aware learning objective that reduces errors stemming from the output-side conditional independence assumption; (3) We introduce a ratio-first decoding strategy that further improve the model's inference efficiency.

\section{Background}
\begin{figure}[t] 
    \centering   
    \setlength{\abovecaptionskip}{3pt}
    \includegraphics[width=0.45\textwidth]{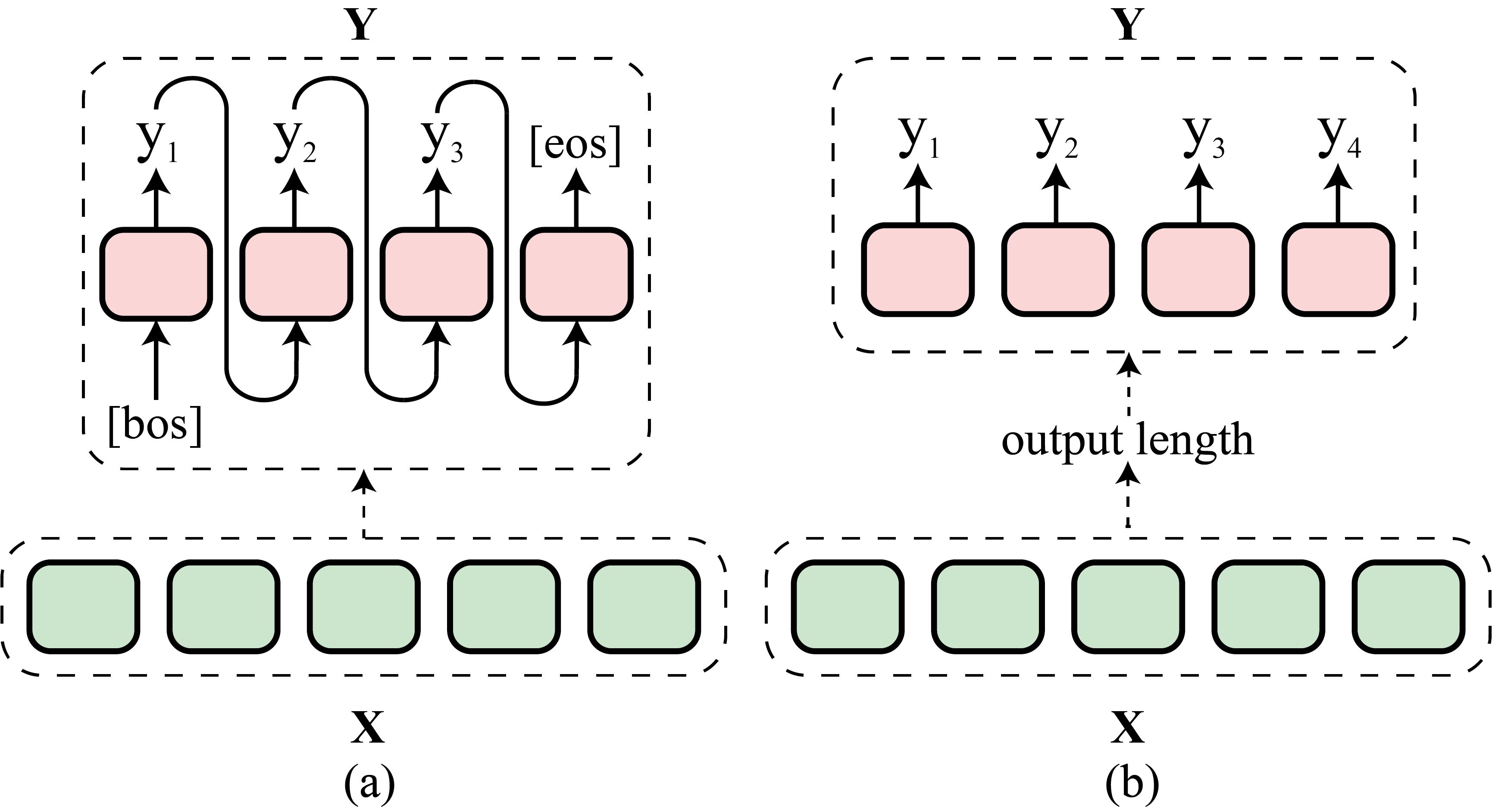}
    \caption{(a) Autoregressive; (b) Non-Autoregressive}
    \label{fig:two-generation}
\end{figure}

Autoregressive generation (AG) models generate sequences based on a left-to-right factorization. As shown in Figure \ref{fig:two-generation}, given the source sequence ${\bf X}$, the target sequence ${\bf Y}$ with length $T^{\prime}$ is generated via a chain of conditional probabilities based on the left-to-right sequential dependencies as:
\begin{equation}
    p({\bf Y}|{\bf X}) = \prod_{i=1}^{T^{\prime}}p(y_i|y_{<i},{\bf X}),
\end{equation}
where $y_{<i}$ denotes the tokens before the $i$-th step. This property of autoregressive factorization makes the generation process hard to be parallelized as the result is generated token by token. 

\begin{figure*}[t] 
	\centering
	\setlength{\abovecaptionskip}{3pt}
	\includegraphics[width=0.95\textwidth]{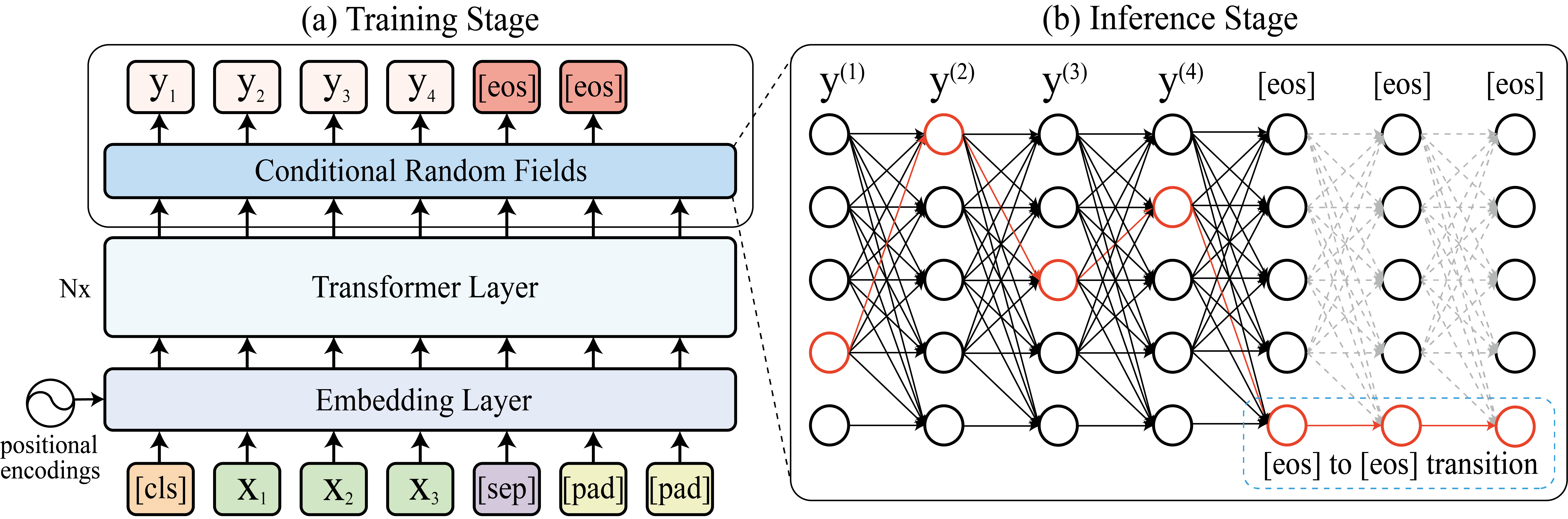}
	\caption{The overall illustration of the proposed model: During training, the model parameters are only updated on the positions of the target sequence. During inference, once the decoded trajectory (colored in \textcolor{red}{red}) gets into the \texttt{[eos]} state, it will only transit to the \texttt{[eos]} state in the remaining steps. The final result is obtained by removing the generated \texttt{[eos]} tokens from the entire decoded trajectory.}
	\label{fig:model}
\end{figure*}

Unlike AG models, non-autoregressive (NAG) models generate sequences without modelling the output-side dependencies. As shown in Figure \ref{fig:two-generation}, given the prespecified output length $T^{\prime}$, the probability of the target sequence ${\bf Y}$ is then modelled as:
\begin{equation}
    p({\bf Y}|{\bf X}) = \prod_{i=1}^{T^{\prime}}p(y_i|{\bf X}, i, T^{\prime}).
\end{equation}
With this conditional independence assumption, NAG models can fully parallelize their generation process, which significantly improves the inference speed. However, it has been shown that, the choice of the prespecified output length has a notable impact on the model's generation quality \cite{DBLP:journals/corr/abs-1711-02281}. In addition, the removal of output-side sequential dependency also causes the generation quality of NAG models to be inferior to their autoregressive counterparts \cite{DBLP:conf/aaai/WangTHQZL19}.

\section{Proposed Model}
In this section, we give a detailed explanation of the proposed model. First, we describe how to utilize BERT as a non-autoregressive generation model. Then we discuss the decoding mechanism which allows the model to determine the output length dynamically. Finally, we introduce the new ratio-first decoding strategy which further improves the model's decoding efficiency. 

\subsection{Model Architecture}
\label{sec:crf}
The architecture of the proposed model is presented in Figure \ref{fig:model}, in which the embedding layer and the stack of transformer layers are initialized with BERT \cite{DBLP:conf/naacl/DevlinCLT19}. 

\paragraph{Input Representation} Following the setup of BERT, we first append a \texttt{[cls]} and a \texttt{[sep]} token on both sides of the source sequence. Then we attach a number of \texttt{[pad]} tokens at the end of source sequence to make its length equal to the predefined maximum size (e.g., 256). Thus we can make sure the source length is longer than or equal to the output length.  As a special case, for tasks like text summarization where the source is known to be longer than the target, we do not attach the \texttt{[pad]} tokens when constructing the input. 

\paragraph{Transformer Layers} Given the source sequence ${\bf X}$, it is processed by a stack of $N$ transformer \cite{DBLP:conf/nips/VaswaniSPUJGKP17} layers. Formally, the Multi-Head Attention is defined as $\text{MultiHead}({\bf Q}, {\bf K}, {\bf V})$, where ${\bf Q}$, ${\bf K}$, ${\bf V}$ denotes the query, key and value respectively. The computation of the first transformer layer is then defined as:
\begin{gather}
{\bf V}^{(1)} = \text{MultiHead}(E({\bf X}), E({\bf X}), E({\bf X})), \\
{\bf O}^{(1)} = \text{FFN}({\bf V}^{(1)}),\\
\text{FFN}(x) = \max(0, xW_1 + b_1)W_2 + b_2,
\end{gather}  
where $E({\bf X})= TE({\bf X}) + PE({\bf X})$ in which $TE(\cdot)$ denotes the token embedding and $PE(\cdot)$ denotes the position embedding. For other layers:
\begin{gather}
{\bf V}^{(n)} = \text{MultiHead}({\bf O}^{(n-1)}, {\bf O}^{(n-1)}, {\bf O}^{(n-1)}), \\
{\bf O}^{(n)} = \text{FFN}({\bf V}^{(n)}),
\end{gather}  
where $n=2,...,N$ and $N$ is the total number of transformer layers. The final sequence representation ${\bf H}\in\mathbb{R}^{T\times d_{\textup{model}}}$ is the output states of BERT from the last layer, where $T$ is the source sequence length and $d_{\textup{model}}$ is the model size.

\paragraph{CRF Layer} Then, ${\bf H}$ is passed through a linear-chain CRF \cite{DBLP:conf/icml/LaffertyMP01}. Under the CRF framework, the likelihood of the target sequence ${\bf Y}$ with length $T^{\prime}$ is then modelled as:
\begin{align}
\begin{split}
\label{score_function}
    &P_{\textup{CRF}}({\bf Y}|{\bf X})= \frac{e^{S({\bf X}, {\bf Y} )}}{\sum_{{\bf Y}^{\prime}}e^{S({\bf X}, {\bf Y}^{\prime})}}\\
    &=\frac{1}{Z({\bf X})}\exp(\sum_{i=1}^{T^{\prime}}\Phi_{y_i}(h_i) + \sum_{i=2}^{T^{\prime}}t(y_{i-1},y_i)),\\
\end{split}
\end{align}
where $Z({\bf X})$ is the normalizing factor and $\Phi_{y_i}(h_i)$ denotes the label score of $y_i$ at position $i$. In practice, $\Phi$ is parameterized by a neural network that maps the BERT output state $h_i$ into the label (vocabulary) space. The $t(y_{i-1},y_i)={\bf T}_{y_{i-1},y_i}$ denotes the transition score from label $y_{i-1}$ to $y_i$ where ${\bf T}\in\mathbb{R}^{|V|\times |V|}$ is the transition matrix. 

\paragraph{Approximation} In the context of text generation, the size of the label space (vocabulary size) $|V|$ is typically large, e.g., $32k$. Therefore, it is intractable to directly model the transition matrix ${\bf T}$ and the normalizing factor $Z({\bf X})$. To this end, we adopt the techniques proposed by \citet{DBLP:conf/nips/SunLWHLD19} to approximate these two terms. Specifically, the full transition matrix is approximated by the product of two low-rank matrices ${\bf T} = {\bf E}_1{\bf E}_2^{T}$, where ${\bf E}_1,{\bf E}_2\in\mathbb{R}^{|V|\times d}$ and $d$ is much smaller than $|V|$. To compute the normalizing factor $Z({\bf X})$, at each time step, instead of searching through all possible paths, the number of candidates is heuristically truncated to a predefined beam size $k$. We refer readers to the original paper for further details. 

\subsection{Output Length Determination}
\label{sec:length_determination}
In this section, we describe how to let the model determine the output sequence length by itself. Our basic idea is that we want the model to dynamically stop generation via emitting a special \texttt{[eos]} token. To achieve this, during training, we manually append \textbf{two} consecutive \texttt{[eos]} tokens to the end of the target sequence, as shown in the top left part of Figure \ref{fig:model}. In this way, the model can learn a deterministic transition behaviour between two \texttt{[eos]} states, meaning that $t(\texttt{[eos]},\texttt{[eos]})=\max_{v\in\mathcal{V}}t(\texttt{[eos]},v)$. This is because, during training, the model never sees a transition (\texttt{[eos]}, $v$), where $v\neq\texttt{[eos]}$.

During inference, the result $\Tilde{{\bf Y}}$ is acquired as $\Tilde{{\bf Y}} = \argmax_{{\bf Y}^{\prime}}S({\bf X}, {\bf Y}^{\prime})$,
where the CRF scoring function $S({\bf X}, {\bf Y}^{\prime})$ in Equation \eqref{score_function} can be decomposed as:
\begin{align}
\begin{split}
    &S({\bf X}, {\bf Y}^{\prime})=\sum_{i=1}^{T}\Phi_{y^{\prime}_i}(h_i) + \sum_{i=2}^{T}t(y^{\prime}_{i-1},y^{\prime}_i)\\
    &=\underbrace{\Phi_{y^{\prime}_1}(h_1)}_{\text{initial state}} + \sum_{i=2}^{T}\{\underbrace{\overbrace{\Phi_{y^{\prime}_i}(h_i)}^{\text{label score}} + \overbrace{t(y^{\prime}_{i-1},y^{\prime}_i)}^{\text{transition score}}}_{\text{state transition}}\}.\\
\end{split}
\end{align}

Once the decoded trajectory enters the \texttt{[eos]} state, the state transition term in $S({\bf X}, {\bf Y}^{\prime})$ will be dominated by the transition score term $t(\texttt{[eos]},\texttt{[eos]})$. As a result, the model will keep transitioning to \texttt{[eos]} in the remaining steps. An example is provided in the right part of Figure \ref{fig:model}, from which we can see that, at step $5$, the decoded trajectory enters the \texttt{[eos]} state and remains at it in the rest of the generation process. In this way, our model can dynamically control the length of output sequence by entering the \texttt{[eos]} state during the generation process. After the entire generation process is completed, the final output sequence can be obtained by removing all generated \texttt{[eos]} tokens.

\subsection{Ratio-First Decoding}
\label{sec:ratio-first}
We note that the outputs of BERT can be divided into two subsets. The first subset ranges from the beginning to the position where the first \texttt{[eos]} is emitted, and the second subset is the rest. For example, in Figure \ref{fig:model}, the first subset are those corresponding to the output sequence ``\texttt{$y^{(1)}$} \texttt{$y^{(2)}$} \texttt{$y^{(3)}$} \texttt{$y^{(4)}$} \texttt{[eos]}”. As for the second part, we can see that it has little effect on the final output and removing it should not change the result. This indicates that it suffices to only consider the beginning part of BERT outputs for improving the inference speed. Especially, for tasks like summarization where the target is known to be shorter than the source sequence, we are safe to only use the first $[\alpha\cdot T]$ outputs of BERT to perform inference. Here $T$ denotes the source length, $\alpha\in(0.0, 1.0)$ is set based on the data statistics and $[\cdot]$ is the integer rounding operation. Formally, given the source sequence ${\bf X}$, the ratio-first decoding is defined as

\begin{align}
\begin{split}
\label{ratio_first}
    &\Tilde{\bf Y} = \argmax_{{\bf Y}^{\prime}}\mathcal{F}({\bf X}, {\bf Y}^{\prime}, \alpha),\\
    &=\argmax_{{\bf Y}^{\prime}}\{\sum_{i=1}^{[\alpha\cdot T]}\Phi_{y^{\prime}_i}(h_i) + \sum_{i=2}^{[\alpha\cdot T]}t(y^{\prime}_{i-1},y^{\prime}_i)\}. 
\end{split}
\end{align}
When $\alpha=1.0$, ratio-first degenerates to the standard decoding strategy in CRF-based models. 

It should be noted that, $[\alpha\cdot T]$ only constrains the maximum length of the generated result, and the actual output length (after removing the generated \texttt{[eos]} tokens) is still decided by the model itself. In the experiment section, we demonstrate that  
ratio-first can notably improve the inference speed whilst maintaining the generation quality.

\section{Learning}
\label{sec:ca-objective}
Due to the conditional independence approximation on output tokens, NAG models often tend to generate repeated tokens \cite{DBLP:conf/aaai/WangTHQZL19}. One way to alleviate this problem is to introduce implicit dependencies on the output side. In this work, we propose to use the unlikelihood formulation of \citet{DBLP:conf/iclr/WelleckKRDCW20} in the context of NAG, where we define the set of negative candidate as the surrounding tokens within a predefined context window $c$. Formally, given the source sequence ${\bf X}$ and the target sequence ${\bf Y}$ with length $T^{\prime}$, the proposed context-aware objective is defined as:
\begin{align}
\begin{split}
\label{ca_loss}
    \mathcal{L}_{\textup{CA}}({\bf Y}|{\bf X}) &= -\sum_{i=1}^{T^{\prime}}\{\log p_{\theta}(y_i|h_i;{\bf X}) + l_{\textup{CA}}(i)\},\\
    l_{\textup{CA}}(i) = &\sum_{j=i-c,y_j\neq y_i}^{j=i+c}\log(1.0-p_{\theta}(y_j|h_i;{\bf X})),\\
\end{split}
\end{align}
where $h_i$ is the model output state at position $i$. At position $i$, the proposed objective maximizes the probability of token $y_i$ while minimizing the probabilities of the surrounding tokens. In this way, it discourages the model from generating repetitive tokens at different time steps.

The overall learning objective is then defined as
\begin{align}
\begin{split}
    \mathcal{L}_{\textup{CRF}} &= -\log P_{\textup{CRF}}({\bf Y}|{\bf X}),\\
    \mathcal{L} &= \mathcal{L}_{\textup{CRF}} + \lambda\cdot\mathcal{L}_{\textup{CA}},\\
\end{split}
\end{align}
where $\lambda$ controls the importance of different loss terms and $P_{\textup{CRF}}({\bf Y}|{\bf X})$ is described in Equation \eqref{score_function}.

\section{Related Work}
Non-Autoregressive generation was first introduced by \citet{DBLP:journals/corr/abs-1711-02281} to reduce the inference latency in machine translation. Recent works in this area have investigated ways to mitigate the trade-off between the decoding speed and generation quality. \citet{DBLP:journals/corr/abs-1711-02281} utilized fertility as latent variables for better translation performance. \citet{DBLP:conf/aaai/WangTHQZL19} proposed two auxiliary objectives for better modelling the output states and solving the under-translation problem. To better model the intermediate alignments between source and target sides, \citet{DBLP:conf/emnlp/MaZLNH19} proposed a model based on the generative flow framework. \citet{DBLP:conf/emnlp/GhazvininejadLL19} proposed to use a masked language objective to train the NAG model. During inference, starting from a fully masked sequence, the output is generated in an iterative refinement manner. Recently, \citet{DBLP:conf/nips/SunLWHLD19} proposed to incorporate a conditional random field into the decoder of a NAG model for better modelling the output-side dependencies. Our work is different from prior works in two aspects: 
(1) we directly utilize a pre-trained language model (BERT) to perform non-autoregressive generation; (2) our model can dynamically generate the output sequence without the need of prespecified output length.

\section{Experiments}
We evaluate the proposed model on three typical text generation tasks: (1) text summarization; (2) sentence compression and (3) machine translation.
\subsection{Experimental Setup}
We implement the proposed model with PyTorch \cite{paszke2017automatic}. The BERT model we use is the Huggingface implementation \cite{Wolf2019HuggingFacesTS} (bert-base-uncased).
To approximate the transition matrix in the CRF layer, we set the dimension $d$ of matrices ${\bf E}_1$ and ${\bf E}_2$ as 32. For the normalizing factor ${\bf Z}({\bf X})$, we set the predefined beam size $k$ as 256. As for the overall learning objective, we set the window size $c$ as $3$ and $\lambda$ as $1.0$. In training, we use Adam optimizer \cite{DBLP:journals/corr/KingmaB14}. To measure the relative speedup, we follow the standard setup which runs inference for each individual example separately. The model's inference speed is computed by averaging the results of test cases. For a fair comparison, we measure the inference speed of all models on the same platform.

\subsection{Text Summarization}
Text summarization aims to automatically generate a compact summary that retains the most important content of the original text document \cite{DBLP:books/sp/mining2012/NenkovaM12}. In this experiment, we use the Gigawords dataset \cite{DBLP:conf/emnlp/RushCW15} as our benchmark. For evaluation, standard metrics including ROUGE-1 (R-1), ROUGE-2 (R-2) and ROUGE-L (R-L) \cite{Lin:2004} are reported.

We compare our model with several representative and the latest NAG models, including NAG-NMT \cite{DBLP:journals/corr/abs-1711-02281}, NAR-REG \cite{DBLP:conf/aaai/WangTHQZL19} and NAG-CRF \cite{DBLP:conf/nips/SunLWHLD19}. Following previous works, during training, we train a length predictor to predict the output length. During inference, for each NAG baseline, we adopt the length-parallel decoding strategy (LPD-$k$) \cite{DBLP:conf/acl/WeiWZLS19}, that is, generating $k$ results using the top-$k$ possible output length predictions from the length predictor. The results are then re-ranked by a transformer model to get the final ouput. In the experiment, we report the results of different NAG baselines using LPD-$9$ decoding. In addition, to better examine the effect of using BERT in NAG models, we add a \textbf{B}NAG-CRF baseline which adopts the same structure of the NAG-CRF model but using BERT as the encoder. We also compare our model with several strong autoregressive models, which are Luong-NMT \cite{DBLP:conf/emnlp/LuongPM15}, Pointer-Generator \cite{DBLP:conf/acl/SeeLM17}, DRGD \cite{DBLP:conf/emnlp/LiLBW17} and Concept Pointer \cite{DBLP:conf/emnlp/WangGHZ19}. To measure the relative inference speedup, we include transformer as a baseline model.


\begin{table}[tb]
    \small
	\renewcommand{\arraystretch}{1.2}
	\centering  
    \begin{center}
    \scalebox{0.93}{
	\begin{tabular}{c|c|c|c|c}
        \hlinewd{0.75pt}
		\textbf{Models}&\textbf{R-1}&\textbf{R-2}&\textbf{R-L}&\textbf{Speedup}\\
		\hline
		\multicolumn{5}{c}{Autoregressive}\\
		\hline
		Luong-NMT&33.10&14.45&30.71&-\\  
		Pointer-Generator&35.98&15.99&33.33&-\\
		DRGD&36.25&17.61&33.55&-\\
		Concept Pointer&36.62&16.40&33.98&-\\  
		Transformer ($\textup{b}=4$)&35.74&16.97&33.43&1.00$\times$\\
        \hline
		\multicolumn{5}{c}{Non-Autoregressive}\\
		\hline
		NAG-NMT&27.20&8.96&25.58&{\bf 9.31}$\times$\\
		+LPD-9&29.76&10.03&28.04&5.28$\times$\\
		\cdashline{1-5}
		NAR-REG&28.56&9.79&26.83&8.64$\times$\\
		+LPD-9&31.23&11.14&29.55&4.74$\times$\\
		\cdashline{1-5}
		NAG-CRF&30.29&12.61&28.71&8.07$\times$\\
		+LPD-9&32.91&14.31&31.03&4.32$\times$\\
		\cdashline{1-5}
		\textbf{B}NAG-CRF&32.63&14.32&30.82&6.13$\times$\\
		+LPD-9&34.56&16.10&32.76&3.21$\times$\\
		\hline
	    Ours ($\alpha=0.3$)&34.67&16.13&32.81&{\bf 9.31}$\times$\\
	    Ours ($\alpha=1.0$)&{\bf 35.05}&{\bf 16.48}&{\bf 33.28}&6.72$\times$\\
	    \hlinewd{0.75pt}
	\end{tabular}}
	\end{center}
    \caption{Results on Gigawords dataset, where $b$ in the transformer baseline stands for beam search size.}
	\label{tb:gigawords}
\end{table}

The results are shown in Table \ref{tb:gigawords}, from which we can see that, by using length-parallel decoding, the performance of all NAG baselines can be notably improved. However, such procedure significantly increases the inference latency. In contrast, our model can self-determine the output length without any re-ranking process. As shown in the results, our model outperforms the best NAG baseline (with LPD) and achieves performances that are comparable with several strong AG models. 

Comparing the results of \textbf{B}NAG-CRF and NAG-CRF, we can see that incorporating BERT as encoder helps to improve the model performance. Nonetheless, our model still outperforms \textbf{B}NAG-CRF with LPD-9 decoding. This is because the dynamic length decoding mechanism allows our model to generate results with optimal length, leading to stronger model performances.

Finally, we analyze the proposed ratio-first decoding. From the results, we observe a moderate performance drop when using ratio-first ($\alpha=0.3$). It comes from the fact that, for some input documents with length $T$, the reference summary is longer than $[\alpha\cdot T]$. In such cases, ratio-first fails to generate the complete reference summary, leading to the drop of performance. On the other hand, we can see that, ratio-first can notably improve the inference speedup. With $\alpha = 0.3$, our model achieves the highest inference speedup while still outperforms all compared NAG models.

\subsection{Sentence Compression}
Sentence compression aims at compressing a long sentence into a short one by deleting redundant words. In this experiment, we use the Google sentence compression dataset \cite{DBLP:conf/emnlp/FilippovaA13} as our benchmark. For evaluation, we use the standard token-kept-F1 (F1) score. In addition, We also report the results of other standard metrics including ROUGE-1, ROUGE-2 and ROUGE-L.

We compare the proposed model with the same NAG baselines as in the previous experiment. 
We also compare our model with several strong autoregressive models, including Bi-LSTM-Dep \cite{DBLP:conf/emnlp/FilippovaACKV15}, Tagger and Tagger+ILP \cite{DBLP:conf/acl/WangJCOSL17}, HiSAN-Dep and HiSAN \cite{DBLP:conf/naacl/KamigaitoHHN18}. To measure the inference speedup, we include transformer as a baseline model.

\begin{table}[tb]
    \small
	\renewcommand{\arraystretch}{1.2}
	\centering  
    \begin{center}
    \scalebox{0.82}{
	\begin{tabular}{c|c|c|c|c|c}
		\hlinewd{0.75pt}
		\textbf{Models}&\textbf{F1}&\textbf{R-1}&\textbf{R-2}&\textbf{R-L}&\textbf{Speedup}\\
		\hline
        \multicolumn{6}{c}{Autoregressive}\\
		\hline
		Bi-LSTM-Dep&82.3&81.5&74.1&81.3&-\\  
		Tagger&82.8&81.1&72.4&80.9&-\\
		Tagger+ILP&79.0&76.1&64.6&75.8&-\\  
		HiSAN-Dep&82.7&82.1&74.9&81.9&-\\   
        HiSAN &83.2&82.9&75.8&82.7&-\\ 
        Transformer ($\textup{b}=4$)&82.4&82.0&74.6&81.8&1.00$\times$\\
        \hline
        \multicolumn{6}{c}{Non-Autoregressive}\\
		\hline
		NAG-NMT&72.5&72.1&59.9&71.8&\textbf{10.71}$\times$\\
		+LPD-9&73.8&73.6&61.0&73.1&6.09$\times$\\
		\cdashline{1-6}
		NAG-REG&73.7&73.1&61.5&73.0&10.00$\times$\\
		+LPD-9&75.6&75.1&63.4&74.9&5.49$\times$\\
		\cdashline{1-6}
		NAG-CRF&75.1&74.4&66.8&74.2&9.41$\times$\\
		+LPD-9&77.3&76.5&69.0&76.3&5.04$\times$\\
		\cdashline{1-6}
		\textbf{B}NAG-CRF&77.1&76.2&68.9&76.0&7.21$\times$\\
		+LPD-9&79.3&78.5&71.7&78.2&3.91$\times$\\
		\hline
		Ours ($\alpha=0.7$)&79.5&79.0&72.1&78.7&10.00$\times$\\
	    Ours ($\alpha=1.0$)&\textbf{80.7}&\textbf{80.3}&\textbf{73.6}&\textbf{80.1}&8.42$\times$\\
		\hlinewd{0.75pt}
	\end{tabular}}
	\end{center}
    \caption{Results on sentence compression task}
	\label{tb:sentence-compression}
\end{table}

The results are presented in Table \ref{tb:sentence-compression}, from which we see that our model outperforms the best reported NAG baseline (with LPD) in terms of both the generation quality and inference speed. Comparing with the strong autoregressive models, our model can achieve competitive performance with a over $8.42\times$ inference speed up. We also report the results of our model using the ratio-first decoding strategy. By setting $\alpha$ as $0.7$, it achieves a $10.00\times$ inference speedup while still outperforming other compared NAG baselines.

\subsection{Machine Translation}
Machine translation aims at translating text from the source language to the target language. In this task, we use the IWSLT14 German-to-English (DE-EN) dataset as our benchmark. Following previous works, we use the sequence-level knowledge distillation \cite{DBLP:journals/corr/abs-1711-02281} during training. For evaluation, we report results in BLEU scores \cite{DBLP:conf/acl/PapineniRWZ02}. In this experiment, we use the BERT model in German language.

We compare our model with a range of strong NAG models, including NAG-NMT \cite{DBLP:journals/corr/abs-1711-02281}, ENAG-E and ENAG-P \cite{DBLP:conf/aaai/GuoTHQXL19}, NAG-REG \cite{DBLP:conf/aaai/WangTHQZL19}, NAG-CRF \cite{DBLP:conf/nips/SunLWHLD19} and \textbf{B}NAG-CRF. For each NAG baseline, we also report the results using LPD-$9$ decoding. In addition, we compare our model with several strong autoregressive models, including LSTM-based  \cite{DBLP:journals/corr/WuSCLNMKCGMKSJL16}, CNN-based  \cite{DBLP:conf/icml/GehringAGYD17} and transformer model. 

The results are shown in Table \ref{tb:machine-translation}, from which we see that our model outperforms the best NAG baseline (with LPD) in terms of both the generation quality and inference speedup. Additionally, we also report the results using the ratio-first decoding. By setting $\alpha$ as $0.8$, the inference speedup can be further boosted to $13.92\times$ while the generation quality is still higher than the best NAG baseline.

\begin{table}[tb]
    \small
    \renewcommand{\arraystretch}{1.2}
	\centering  
    \begin{center}
    \scalebox{0.98}{
	\begin{tabular}{c|c|c}
	    \hlinewd{0.75pt}
		\textbf{Models}&\textbf{BLEU}&\textbf{Speedup($\times$)}\\
		\hline
		\multicolumn{3}{c}{Autoregressive}\\
		\hline
		LSTM-based&28.53&-\\
		CNN-based&32.84&-\\
		Transformer ($\textup{b}=4$)&33.31&1.00\\
		\hline
		\multicolumn{3}{c}{Non-Autoregressive}\\
		\hline
		ENAG-E&24.13 (27.30)&15.08 (7.39)\\
        ENAG-P&25.09 (28.60)&14.48 (7.24)\\
        NAG-REG&23.89 (28.04)&16.45 (9.05)\\
		NAG-NMT&23.04 (26.79)&13.92 (7.24)\\
		NAG-CRF&26.39 (29.21)&11.74 (6.03)\\
		\textbf{B}NAG-CRF&26.73 (29.67)&9.42 (5.01)\\
		\hline
		Ours ($\alpha=0.8$)&29.71&13.92\\
		Ours ($\alpha=1.0$)&\textbf{30.45}&11.31\\
		\hlinewd{0.75pt}
	\end{tabular}}
	\end{center}
    \caption{Results on IWSLT14 De-En dataset. The numbers in () are results using length-parallel decoding.}
	\label{tb:machine-translation}
\end{table}

\subsection{Further Analysis}
In this section, we present further discussions and empirical analysis of the proposed model.

\begin{table}[tb]
    \small
    \renewcommand{\arraystretch}{1.2}
	\centering  
    \begin{center}
    \scalebox{1.0}{
	\begin{tabular}{c|c|c|c|c}
		\hlinewd{0.75pt}
		\textbf{BERT}&\textbf{CRF}&\textbf{R-1}&\textbf{R-2}&\textbf{R-L}\\
		\hline
        \checkmark&\checkmark&35.05&16.48&33.28\\
        \cdashline{1-5}
        $\times$&\checkmark&32.41&14.19&30.53\\
        \checkmark&$\times$&32.16&11.33&30.34\\
        $\times$&$\times$&27.02&8.81&25.25\\
		\hlinewd{0.75pt}
	\end{tabular}}
	\end{center}
    \caption{Ablation study on Gigawords dataset.}
	\label{tb:gigawords-ablation}
\end{table}

\paragraph{BERT \& CRF}
To quantify the importance of each component (BERT \& CRF) of our model, we evaluate the performance on Gigawords dataset by removing each component iteratively. 

The results are shown in Table \ref{tb:gigawords-ablation}, from which we can see that by removing any of these components, the overall performance decreases. By removing BERT from the model, we observe notable drop across all metrics. This shows that the knowledge of BERT is an important factor of the model's strong performance. Comparing with results in Table \ref{tb:gigawords}, it still outperforms vanilla NAG-CRF and performs comparably with NAG-CRF using LPD decoding, which demonstrates the merit of the proposed dynamic length decoding mechanism. Another interesting finding is that, by only removing the CRF layer, the most notable drop is observed on the bigram-level metric (ROUGE-2). This shows that the bigram-level dependencies on the output side are mainly captured by the CRF module. In addition, by removing both BERT and CRF, all metrics further decrease. This confirms that each of these two components positively contributes to the model's overall performance.

\begin{table}[tb]
    \small
    \renewcommand{\arraystretch}{1.2}
	\centering 
    \begin{center}
    \scalebox{0.95}{
	\begin{tabular}{c|c|c|c|c|c}
		\hlinewd{0.75pt}
		\textbf{Models}&\textbf{rep-$1$}&\textbf{rep-$2$}&\textbf{rep-$3$}&\textbf{rep-$4$}&\textbf{R-L}\\
		\hline
		w/o CA&6.897&2.640&0.741&0.295&32.89\\
		Ours &5.786&1.978&0.427&0.106&33.28\\
        \cdashline{1-6}
        Transformer &4.329&1.348&0.267&0.089&33.43\\
		\hlinewd{0.75pt}
	\end{tabular}}
	\end{center}
    \caption{Evaluation results on $n$-gram repetitions.}
	\label{tb:rep}
\end{table}

\paragraph{Context-Aware Objective}
In this part, we study the effect of the context-aware objective. As described in Equation \eqref{ca_loss}, it aims at alleviating the problem of repetitive generation. To give a quantitative analysis, we use the measurement of sentence-level repetition \cite{DBLP:conf/iclr/WelleckKRDCW20} to compute the ratio of duplicate $n$-grams (rep-$n$) in the generated result. This metric is defined as 
\begin{equation}
    \text{rep-}n({\bf Y}) = 100\times(1.0 - \frac{|\text{unique } n\text{-grams}({\bf Y})|}{|n\text{-grams}({\bf Y})|}).
\end{equation}
For each generated result, rep-$n$ is $0.0$ when it has no repeating $n$-grams. The final result is computed by averaging over the entire evaluation set.

We conduct experiments on Gigawords dataset to evaluate the $n$-gram repetitions ranging from uni-gram to $4$-gram. The results are shown in Table \ref{tb:rep}, where w/o CA means the model is trained without using context-aware objective and R-L denotes the model's ROUGE-L score. Additionally, we also show the results from transformer model for a direct comparison. Comparing the two variants of our model, we see that training with context-aware objective leads to a 42\% drop on rep-$3$ metric ($0.427$ vs $0.741$) and a 64\% drop on rep-$4$ metric ($0.106$ vs $0.295$). The ROUGE-L results also indicate that the reduction in token repetition can effectively improve the model generation quality. 

\paragraph{Dynamic Length Determination} Next, we examine the importance of the model's ability to dynamically determine the length of the generated output. To this end, we train another model variant by removing the two \texttt{[eos]} tokens from the target sequence. In this way, the model is not able to self-determine the output length throughout the generation process. To perform inference, we use length-parallel decoding (LPD) with different number of length candidates. Formally, for each length candidate $l$, the model generates the result $\Tilde{\bf Y}$ as
\begin{equation}
    \Tilde{\bf Y}=\argmax_{{\bf Y}^{\prime}}\{\sum_{i=1}^{l}\Phi_{y^{\prime}_i}(h_i) + \sum_{i=2}^{l}t(y^{\prime}_{i-1},y^{\prime}_i)\}.
\end{equation}
The final result is acquired by re-ranking the generated results with a transformer model.

We conduct experiments on the IWSLT14 DE-EN dataset in which we try a different number of length candidates, including top-$1$, top-$5$ and top-$10$. The results are shown in Table \ref{tb:mt_length}, from which we can see, as the number of length candidates increases, the model performance increases as well. The reason is that a larger candidates set is more likely to contain the best-suited length for the generation model, leading to better performance. However, such decoding procedure inevitably increases the required computation overhead. We can see that, when setting $k$ as $10$, the inference speedup decreases from $11.84\times$ to $6.01\times$. In contrast, our proposed model is able to determine the optimal output length by itself. Without any re-ranking process, it outperforms the model with LPD-$10$ decoding and achieves the inference speedup that is comparable with the model using LPD-$1$ decoding.

\begin{table}[tb]
    \small
    \renewcommand{\arraystretch}{1.2}
	\centering  
    \begin{center}
    \scalebox{0.9}{
	\begin{tabular}{c|c|c|c|c}
		\hlinewd{0.75pt}
		\multirow{2}{*}{\textbf{Models}}&\multirow{1}{*}{Ours}&\multicolumn{3}{c}{Length-Parallel Decoding}\\\cline{3-5}
		&($\alpha=1.0$)&LPD-$1$&LPD-$5$&LPD-$10$\\
		\hline
        BLEU&\textbf{30.45}&27.15&29.62&30.37\\
        \hline
        Speedup($\times$)&11.31&11.84&8.92&6.01\\
		\hlinewd{0.75pt}
	\end{tabular}}
	\end{center}
    \caption{Results comparison on IWSLT14 dataset}
	\label{tb:mt_length}
\end{table}

\begin{figure}[t] 
	\centering    
	\setlength{\abovecaptionskip}{3pt}
	\includegraphics[width=0.48\textwidth]{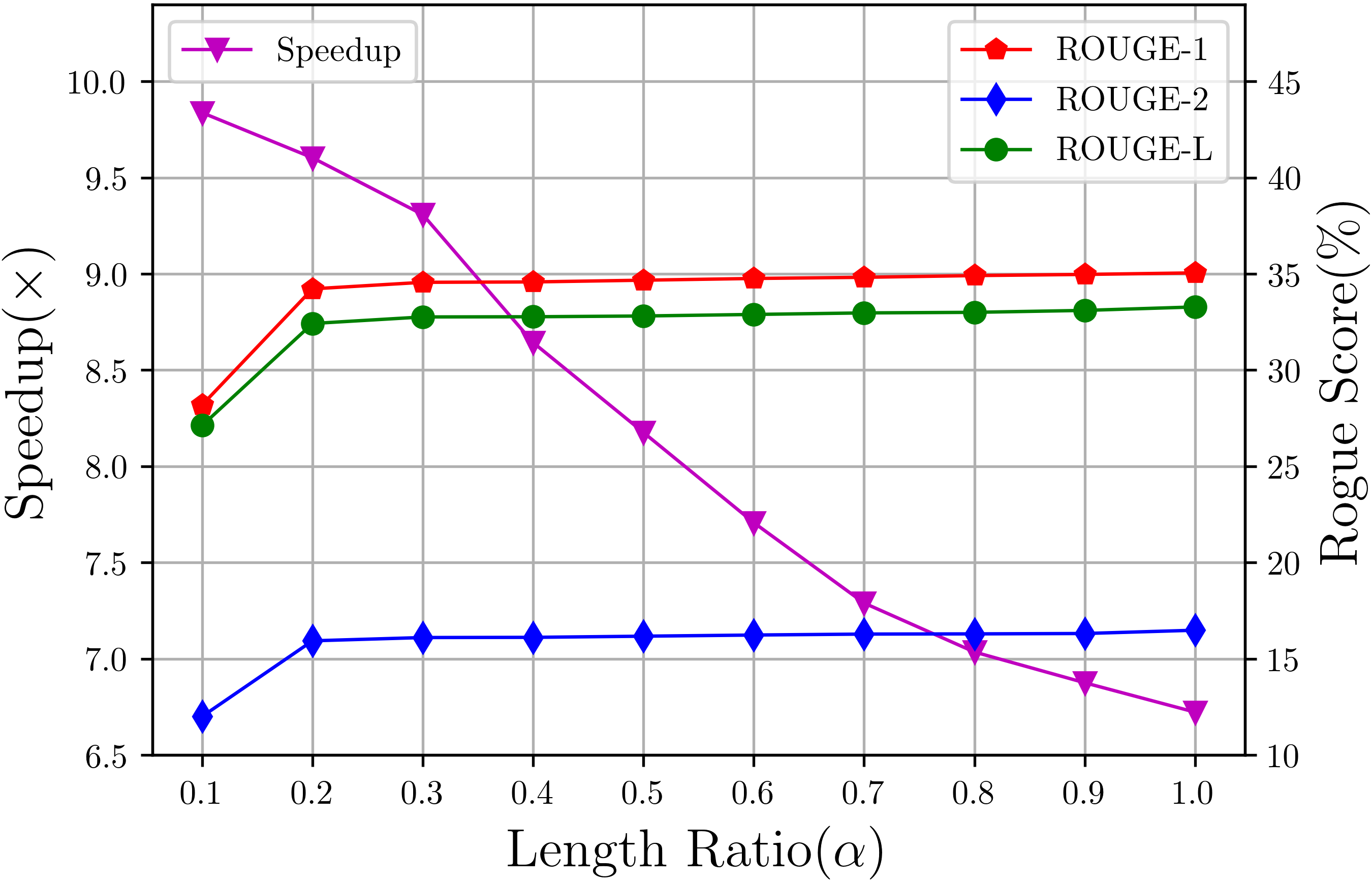}
	\caption{Experiment results on Gigawords dataset using ratio-first decoding with different $\alpha$.}
	\label{fig:performance}
\end{figure}

\paragraph{Ratio-First Decoding}
We are also interested in the effect of the ratio-first decoding strategy. To provide a quantitative analysis, we perform inference on the Gigawords dataset using ratio-first with different $\alpha$. The experimental results with different $\alpha$ are presented in Figure \ref{fig:performance}. It can be observed that, when $\alpha$ reaches $0.3$, the model approximately achieves its optimal performance. At the same time, a notable improvement can be observed in terms of the inference speedup ($6.72\times\rightarrow 9.31\times$). 

\begin{figure}[t] 
	\centering    
	\setlength{\abovecaptionskip}{3pt}
	\includegraphics[width=0.43\textwidth]{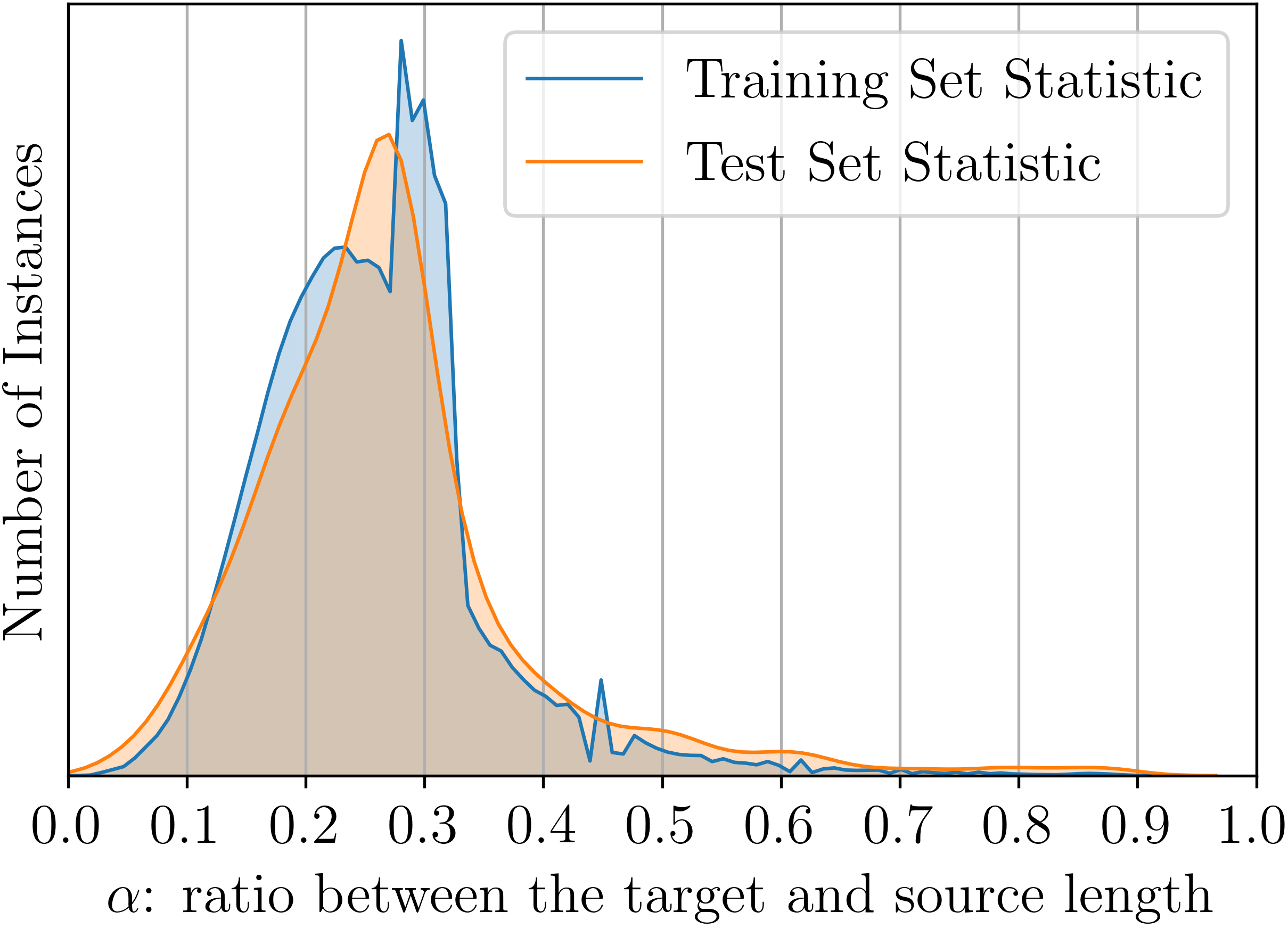}
	\caption{The distribution of target/source length ratio of the training and test set in Gigawords dataset.}
	\label{fig:statistic}
\end{figure}

Now we illustrate why the near optimal performance can be achieved when $\alpha$ reaches $0.3$. In Figure \ref{fig:statistic}, we present the distribution of the target/source length ratio of every data instance in the Gigawords dataset. We can see that, for most cases, the ratio between the target length $T^{\prime}$ and source length $T$ is less than $0.3$. Recall the definition of ratio-first decoding in Equation \eqref{ratio_first}, the $[\alpha\cdot T]$ constrains the maximum length of the generated result. Therefore, once we have a prior knowledge on the data statistic, we can easily choose a proper $\alpha$ that both improves the inference speed whilst maintaining the generation quality. In this case, a proper $\alpha$ could be $0.3$ which is demonstrated by the results in Figure \ref{fig:performance} and \ref{fig:statistic}. By setting different $\alpha$, ratio-first provides us an explicit way to control the balance between the inference speed and the generation quality. This property of ratio-first is especially favorable in real-life scenarios where the inference speed is the highest concern. 

\section{Conclusion}
In this work, we explored the potential of BERT in various text generation tasks under the NAG framework. To address problems from NAG models previously having a prefixed output length, we devised a decoding mechanism which enables the model to determine the output length dynamically. To reduce errors stemming from the assumption of conditional independence of output tokens, we proposed a context-aware objective as well as using a CRF decoding. Furthermore, to maximize the inference speed advantage of our model, we introduced a ratio-first decoding strategy. We evaluated our model on three benchmark datasets and the results show that our model significantly outperforms many strong NAG baselines and performs comparably to many strong AG models.

\section*{Acknowledgments}
The authors wish to thank Jialu Xu, Guanlin Li, Xing Wang for their insightful discussions and support. Many thanks to our anonymous reviewers for their suggestions and comments.

\bibliographystyle{acl_natbib}
\bibliography{anthology,eacl2021}

\end{document}